\def\year{2018}\relax
\documentclass[letterpaper]{article} 
\usepackage{aaai18} 
\usepackage{times} 
\usepackage{helvet} 
\usepackage{courier} 
\usepackage{url} 
\usepackage{graphicx} 
\begin{document}
%
\title{Augmenting End-to-End Dialogue Systems with Commonsense Knowledge}
\author{Tom Young$^1$, Erik Cambria$^2$, Iti Chaturvedi$^2$, Hao Zhou$^3$, Subham Biswas$^2$, Minlie Huang$^3$\\
$^1$School of Information and Electronics, Beijing Institute of Technology, China\\
$^2$School of Computer Science and Engineering, Nanyang Technological University, Singapore\\
$^3$Department of Computer Science and Technology, Tsinghua University, China\\
\textsc{tom@sentic.net, cambria@ntu.edu.sg, iti@ntu.edu.sg}\\ \textsc{tuxchow@gmail.com, subham@sentic.net, aihuang@tsinghua.edu.cn}}

\maketitle

\begin{abstract}

Building dialogue systems that can converse naturally with humans is a challenging yet intriguing problem of artificial intelligence. In open-domain human-computer conversation, where the conversational agent is expected to respond to human utterances in an interesting and engaging way, commonsense knowledge has to be integrated into the model effectively. In this paper, we investigate the impact of providing commonsense knowledge about the concepts covered in the dialogue. Our model represents the first attempt to integrating a large commonsense knowledge base into end-to-end conversational models. In the retrieval-based scenario, we propose a model to jointly take into account message content and related commonsense for selecting an appropriate response. Our experiments suggest that the knowledge-augmented models are superior to their knowledge-free counterparts.

\end{abstract}

\section{Introduction}\label{sec:introduction}
In recent years, data-driven approaches to building conversation models have been made possible by the proliferation of social media conversation data and the increase of computing power. By relying on a large number of message-response pairs, the Seq2Seq framework~\cite{sutskever2014sequence} attempts to produce an appropriate response based solely on the message itself, without any memory module. 

\begin{figure*}[ht]
 \includegraphics[height=2.6cm]{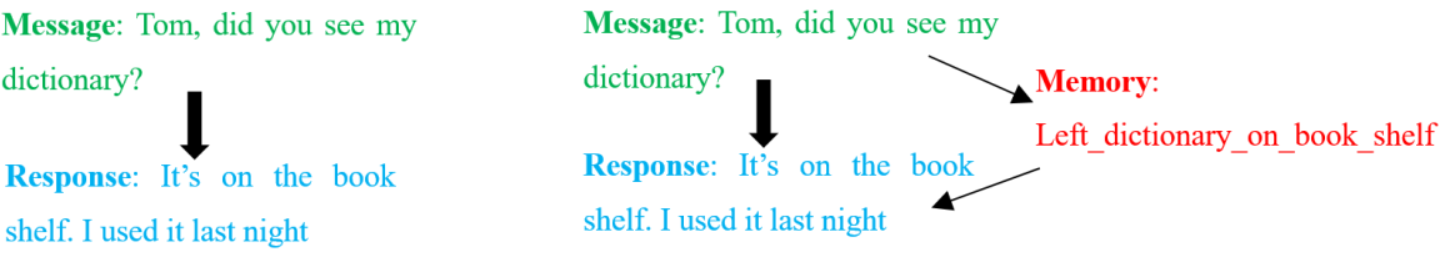}
 \centering
 \caption{Left: In traditional dialogue systems, the response is determined solely by the message itself (arrows denote dependencies). Right: The responder recalls relevant information from memory; memory and message content jointly determine the response. In the illustrated example, the responder retrieves the event ``Left\_dictionary\_on\_book\_shelf'' from memory, which triggers a meaningful response.}
 \label{fig:memory_module}
\end{figure*}

In human-to-human conversations, however, people respond to each other's utterances in a meaningful way not only by paying attention to the latest utterance of the conversational partner itself, but also by recalling relevant information about the concepts covered in the dialogue and integrating it into their responses. Such information may contain personal experience, recent events, commonsense knowledge and more (Figure~\ref{fig:memory_module}). As a result, it is speculated that a conversational model with a ``memory look-up'' module can mimic human conversations more closely~\cite{DBLP:journals/corr/GhazvininejadBC17,bordes2016learning}.
In open-domain human-computer conversation, where the model is expected to respond to human utterances in an interesting and engaging way, commonsense knowledge has to be integrated into the model effectively. 

In the context of artificial intelligence (AI), commonsense knowledge is the set of background information that an individual is intended to know or assume and the ability to use it when appropriate~\cite{minsoc,camcom,camsen}. Due to the vastness of such kind of knowledge, we speculate that this goal is better suited by employing an external memory module containing commonsense knowledge rather than forcing the system to encode it in model parameters as in traditional methods. 

In this paper, we investigate how to improve end-to-end dialogue systems by augmenting them with commonsense knowledge, integrated in the form of external memory. The remainder of this paper is as follows: next section proposes related work in the context of conversational models and commonsense knowledge; following, a section describes the proposed model in detail; later, a section illustrates experimental results; finally, the last section proposes concluding remarks and future work.

\section{Related Work}\label{sec:related_work}
\label{gen_inst}

\subsection{Conversational Models}\label{sec:conversational_models}

Data-driven conversational models generally fall into two categories: retrieval-based methods~\cite{lowe2015ubuntu,lowe2016evaluation,zhou2016multi}, which select a response from a predefined repository, and generation-based methods~\cite{ritter2011data,serban2016building,vinyals2015neural}, which employ an encoder-decoder framework where the message is encoded into a vector representation and, then, fed to the decoder to generate the response. The latter is more natural (as it does not require a response repository) yet suffers from generating dull or vague responses and generally needs a great amount of training data.

The use of an external memory module in natural language processing (NLP) tasks has received considerable attention recently, such as in question answering~\cite{weston2015towards} and language modeling~\cite{sukhbaatar2015end}. It has also been employed in dialogue modeling in several limited settings. With memory networks, \cite{dodge2015evaluating} used a set of fact triples about movies as long-term memory when modeling reddit dialogues, movie recommendation and factoid question answering. 
Similarly in a restaurant reservation setting, \cite{bordes2016learning} provided local restaurant information to the conversational model. 

Researchers have also proposed several methods to incorporate knowledge as external memory into the Seq2Seq framework. \cite{DBLP:journals/corr/XingWWLHZM16} incorporated the topic words of the message obtained from a pre-trained latent Dirichlet allocation (LDA) model into the context vector through a joint attention mechanism. \cite{DBLP:journals/corr/GhazvininejadBC17} mined FoodSquare tips to be searched by an input message in the food domain and encoded such tips into the context vector through one-turn hop. The model we propose in this work shares similarities with~\cite{Lowe-unstructured-text-2015-nips-workshop}, which encoded unstructured textual knowledge with a recurrent neural network (RNN). Our work distinguishes itself from previous research in that we consider a large heterogeneous commonsense knowledge base in an open-domain retrieval-based dialogue setting.

\subsection{Commonsense Knowledge}\label{sec:commonsense_knowledge}

\begin{figure}[h]
 \includegraphics[width=\linewidth]{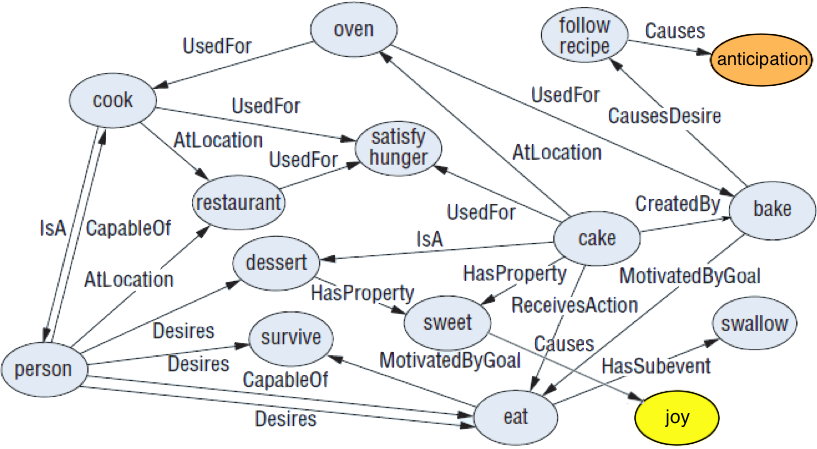}
 \centering
 \caption{A sketch of SenticNet semantic network.}
 \label{fig:affnet}
\end{figure}

Several commonsense knowledge bases have been constructed during the past decade, such as ConceptNet~\cite{specon} and SenticNet~\cite{camnt4}. The aim of commonsense knowledge representation and reasoning is to give a foundation of real-world knowledge to a variety of AI applications, e.g., sentiment analysis~\cite{porsent}, handwriting recognition~\cite{wancom}, e-health~\cite{campat}, aspect extraction~\cite{porlda}, and many more. Typically, a commonsense knowledge base can be seen as a \emph{semantic network} where \emph{concepts} are nodes in the graph and \emph{relations} are edges (Figure~\ref{fig:affnet}). Each $\textless concept1, relation, concept2 \textgreater$ triple is termed an \emph{assertion}. 

Based on the Open Mind Common Sense project~\cite{singh2002open}, ConceptNet not only contains objective facts such as ``Paris is the capital of France'' that are constantly true, but also captures informal relations between common concepts that are part of everyday knowledge such as ``A dog is a pet''. This feature of ConceptNet is desirable in our experiments, because the ability to recognize the informal relations between common concepts is necessary in the open-domain conversation setting we are considering in this paper.

\section{Model Description}\label{sec:model_description}

\label{headings}

\subsection{Task Definition}\label{sec:task_definition}

In this work, we concentrate on integrating commonsense knowledge into retrieval-based conversational models, because they are easier to evaluate~\cite{liu2016not,lowe2016evaluation} and generally take a lot less data to train. We leave the generation-based scenario to future work.

\emph{Message} (\emph{context}) $x$ and \emph{response} $y$ are a sequence of tokens from vocabulary $V$. Given $x$ and a set of response candidates $[y_1,y_2,y_3...,y_K]\in Y$, the model chooses the most appropriate response $\hat{y}$ according to:
\begin{eqnarray}
\hat{y}=\mathop{\arg\max}_{y\in{Y}}f(x,y),
\label{eq1}
\end{eqnarray}

%

where $f(x,y)$ is a scoring function measuring the ``compatibility'' of $x$ and $y$. 
The model is trained on $\textless message, response, label \textgreater$ triples with cross entropy loss, where $label$ is binary indicating whether the $\textless message, response \textgreater$ pair comes from real data or is randomly combined. 

\subsection{Dual-LSTM Encoder}\label{duel_LSTM_encoder}

As a variation of vanilla RNN, a long short-term memory (LSTM) network~\cite{hochreiter1997long} is good at handling long-term dependencies and can be used to map an utterance to its last hidden state as fixed-size embedding representation. 
The Dual-LSTM encoder~\cite{lowe2015ubuntu} represents the message $x$ and response $y$ as fixed-size embeddings $\vec{x}$ and $\vec{y}$ with the last hidden states of the same LSTM. The compatibility function of the two is thus defined by:
\begin{eqnarray}
f(x,y) = \sigma(\vec{x}^{T}W\vec{y}),
\end{eqnarray}

where matrix $W \in \mathcal{R}^{D\times D}$ is learned during training.

\begin{figure*}[ht]
 \includegraphics[height=10.5cm]{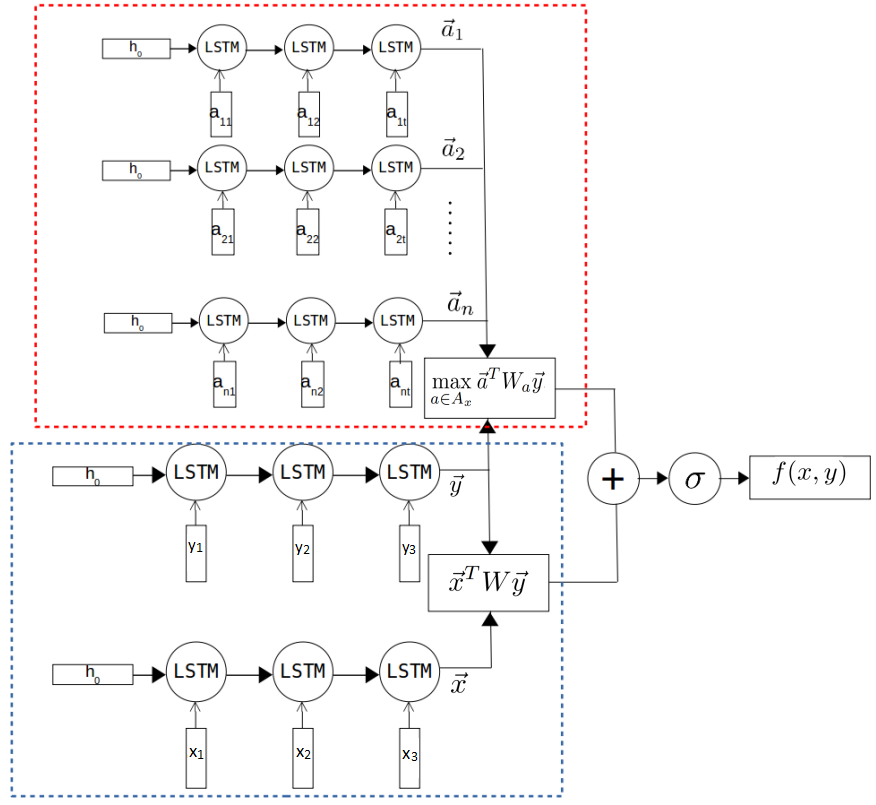}
 \centering
 \caption{Tri-LSTM encoder. We use LSTM to encode message, response and commonsense assertions. LSTM weights for message and response are tied. The lower box is equal to a Dual-LSTM encoder. The upper box is the memory module encoding all commonsense assertions.}
 \label{fig:triencoder}
\end{figure*}

\subsection{Commonsense Knowledge Retrieval}\label{sec:commonsense_knowledge_retrieval}

In this paper, we assume that a commonsense knowledge base is composed of assertions $A$ about concepts $C$. Each assertion $a \in A$ takes the form of a triple $\textless c_1,r,c_2 \textgreater$, where $r \in R$ is a \emph{relation} between $c_1$ and $c_2$, such as \emph{IsA}, \emph{CapableOf}, etc. $c_1,c_2$ are concepts in $C$. The relation set $R$ is typically much smaller than $C$. $c$ can either be a single word (e.g., ``dog'' and ``book'') or a multi-word expression (e.g., ``take\_a\_stand'' and ``go\_shopping'').
We build a dictionary $H$ out of $A$ where every concept $c$ is a key and a list of all assertions in $A$ concerning $c$, i.e., $c=c_1$ or $c=c_2$, is the value.
Our goal is to retrieve commonsense knowledge about every concept covered in the message. 

\begin{figure*}[ht]
\includegraphics[height=8cm]{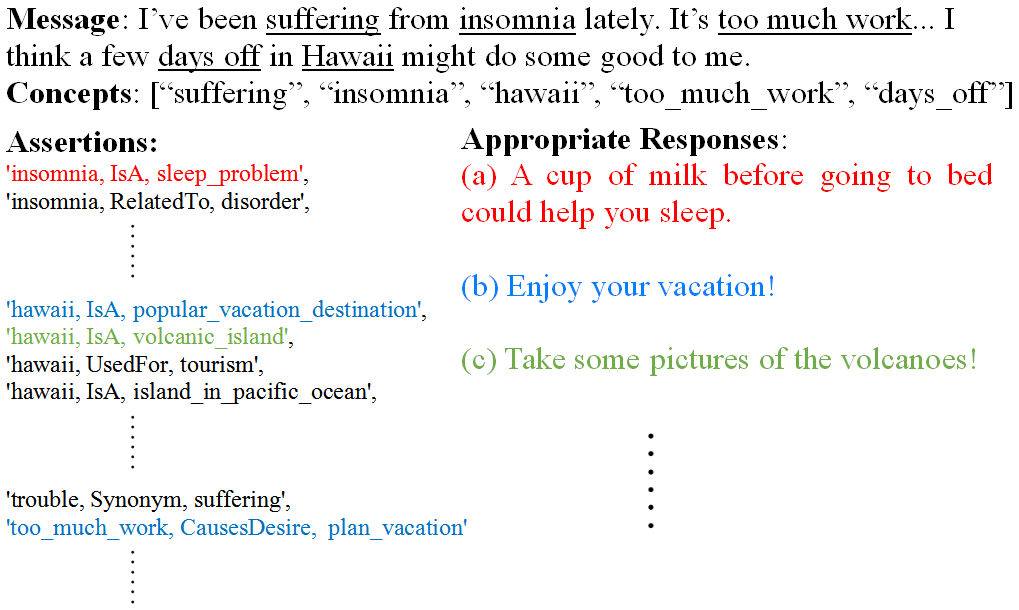}
 \centering
 \caption{In the illustrated case, five concepts are identified in the message. All assertions associated with the five concepts constitute $A_x$. We show three appropriate responses for this single message. Each of them is associated with (same color) only one or two commonsense assertions, which is a paradigm in open-domain conversation and provides ground for our max-pooling strategy. It is also possible that an appropriate response is not relevant to any of the common assertions in $A_x$ at all, in which case our method falls back to Dual-LSTM.}
 \label{fig:instance_csk}
\end{figure*}

We define $A_x$ as the set of commonsense assertions concerned with message $x$. To recover concepts in message $x$, we use simple $n$-gram matching ($n\leq N$)\footnote{More sophisticated methods such as $concept\ parser$~\cite{rajagopal2013graph} are also possible. Here, we chose n-gram for better speed and recall. $N$ is set to 5.}. Every $n$-gram in $c$ is considered a potential concept\footnote{For unigrams, we exclude a set of stopwords. Both the original version and stemmed version of every word are considered.}. If the $n$-gram is a key in $H$, the corresponding value, i.e., all assertions in $A$ concerning the concept, is added to $A_x$ (Figure~\ref{fig:instance_csk}).

\subsection{Tri-LSTM Encoder}\label{sec:tri_lstm_encoder}

Our main approach to integrating commonsense knowledge into the conversational model involves using another LSTM for encoding all assertions $a$ in $A_x$, as illustrated in Figure~\ref{fig:triencoder}. Each $a$, originally in the form of $\textless c_1,r,c_2 \textgreater$, is transformed into a sequence of tokens by chunking $c_1$, $c_2$, concepts which are potentially multi-word phrases, into $[c_{11},c_{12},c_{13}...]$ and $[c_{21},c_{22},c_{23}...]$. Thus, $a=[c_{11},c_{12},c_{13}...,r,c_{21},c_{22},c_{23}...]$. 

We add $R$ to vocabulary $V$, that is, each $r$ in $R$ will be treated like any regular word in $V$ during encoding. We decide not to use each concept $c$ as a unit for encoding $a$ because $C$ is typically too large ($>$1M).
$a$ is encoded as embedding representation $\vec{a}$ using another LSTM. Note that this encoding scheme is suitable for any natural utterances containing commonsense knowledge\footnote{Termed \emph{surface text} in ConceptNet.} in addition to well-structured assertions. We define the \emph{match score} of assertion $a$ and response $y$ as:
\begin{eqnarray}
m(a,y) = \vec{a}^{T}W_a\vec{y},
\end{eqnarray}


where $W_a \in \mathcal{R}^{D\times D}$ is learned during training. 
Commonsense assertions $A_x$ associated with a message is usually large ($>$100 in our experiment). 
We observe that in a lot of cases of open-domain conversation, response $y$ can be seen as triggered by certain perception of message $x$ defined by one or more assertions in $A_x$, as illustrated in Figure~\ref{fig:instance_csk}. We can see the difference between message and response pair when commonsense knowledge is used. For example, the word `Insomnia' in the message is mapped to the commonsense assertion `Insomnia, IsA, sleep$\_$problem'. The appropriate response is then matched to `sleep$\_$problem' that is `go to bed'. Similarly, the word `Hawaii' in the message is mapped to the commonsense assertion `Hawaii, UsedFor, tourism'. The appropriate response is then matched to `tourism' that is `enjoy vacation'. In this way, new words can be mapped to the commonly used vocabulary and improve response accuracy.  

Our assumption is that $A_x$ is helpful in selecting an appropriate response $y$. However, usually very few assertions in $A_x$ are related to a particular response $y$ in the open-domain setting. As a result, we define the \emph{match score} of $A_x$ and $y$ as 
\begin{eqnarray}
m(A_x,y)=\mathop{\max}_{a\in{A_x}} m(a,y),
\end{eqnarray}

%

that is, we only consider the commonsense assertion $a$ with the highest match score with $y$, as most of $A_x$ are not relevant to $y$. Incorporating $m(A_x,y)$ into the Dual-LSTM encoder, our Tri-LSTM encoder model is thus defined as:
\begin{eqnarray}
f(x,y) = \sigma(\vec{x}^{T}W\vec{y} + m(A_x,y)),
\end{eqnarray}
i.e., we use simple addition to supplement $x$ with $A_x$, without introducing a mechanism for any further interaction between $x$ and $A_x$. This simple approach is suitable for response selection and proves effective in practice. 

The intuition we are trying to capture here is that an appropriate response $y$ should not only be compatible with $x$, but also related to certain memory recall triggered by $x$ as captured by $m(A_x,y)$. In our case, the memory is commonsense knowledge about the world. In cases where $A_x = \emptyset$, i.e., no commonsense knowledge is recalled, $m(A_x,y)=0$ and the model degenerates to Dual-LSTM encoder.

\subsection{Comparison Approaches}\label{comparison_approaches}

%

\subsubsection{Supervised Word Embeddings}\label{supervised_word_embeddings}
We follow~\cite{bordes2016learning,dodge2015evaluating} and use supervised word embeddings as a baseline. Word embeddings are most well-known in the context of unsupervised training on raw text as in~\cite{mikolov2013efficient}, yet they can also be used to score message-response pairs. The embedding vectors are trained directly for this goal. In this setting, the ``compatibility'' function of $x$ and $y$ is defined as:
\begin{eqnarray}
f(x,y)=\vec{x}^T\vec{y}
\end{eqnarray}
In this setting, $\vec{x},\vec{y}$ are bag-of-words embeddings.
With retrieved commonsense assertions $A_x$, we embed each $a\in{A_x}$ to bag-of-words representation $\vec{a}$ and have:
\begin{eqnarray}
f(x,y)=\vec{x}^T\vec{y}+\mathop{\max}_{a\in{A_x}} \ \ \vec{a}^T\vec{y}.
\end{eqnarray}
This linear model differs from Tri-LSTM encoder in that it represents an utterance with its bag-of-words embedding instead of RNNs.

\subsubsection{Memory Networks}\label{memory_networks}
Memory networks~\cite{sukhbaatar2015end,weston2014memory} are a class of models that perform language understanding by incorporating a memory component. They perform attention over memory to retrieve all relevant information that may help with the task. In our dialogue modeling setting, we use $A_x$ as the memory component. Our implementation of memory networks, similar to~\cite{bordes2016learning,dodge2015evaluating}, differs from supervised word embeddings described above in only one aspect: how to treat multiple entries in memory.
In memory networks, output memory representation $\vec{o}=\sum_{i}p_i\vec{a}_i$, where $\vec{a}_i$ is the bag-of-words embedding of $a_i\in{A_x}$ and $p_i$ is the attention signal over memory $A_x$ calculated by $p_i=softmax(\vec{x}^T\vec{a_i})$. The ``compatibility'' function of $x$ and $y$ is defined as:
\begin{eqnarray}
f(x,y)=(\vec{x}+\vec{o})^T\vec{y}=\vec{x}^T\vec{y}+(\sum_{i}p_i\vec{a}_i)^T\vec{y}
\end{eqnarray}

In contrast to supervised word embeddings described above, attention over memory is determined by message $x$. This mechanism was originally designed to retrieve information from memory that is relevant to the context, which in our setting is already achieved during commonsense knowledge retrieval. As speculated, the attention over multiple memory entries is better determined by response $y$ in our setting. We empirically prove this point below. 

\section{Experiments}\label{experiments}

\subsection{Twitter Dialogue Dataset}\label{twitter_dialogue_dataset}

To the best of our knowledge, there is currently no well-established open-domain response selection benchmark dataset available, although certain Twitter datasets have been used in the response generation setting~\cite{li2015diversity,li2016persona}. We thus evaluate our method against state-of-the-art approaches in the response selection task on Twitter dialogues.
	
1.4M Twitter \textless message, response$>$ pairs are used for our experiments. They were extracted over a 5-month period, from February through July in 2011. 1M Twitter \textless message, response$>$ pairs are used for training. With the original response as ground truth, we construct 1M \textless message, response, label=1$>$ triples as positive instances. Another 1M negative instances \textless message, response, label=0$>$ are constructed by replacing the ground truth response with a random response in the training set.

For tuning and evaluation, we use 20K \textless message, response$>$ pairs that constitute the validation set (10K) and test set (10K). They are selected by a criterion that encourages interestingness and relevance: both the message and response have to be at least 3 tokens long and contain at least one non-stopword. For every message, at least one concept has to be found in the commonsense knowledge base. For each instance, we collect another 9 random responses from elsewhere to constitute the response candidates. 

Preprocessing of the dataset includes normalizing hashtags, ``@User'', URLs, emoticons. Vocabulary $V$ is built out of the training set with 5 as minimum word frequency, containing 62535 words and an extra $\textless UNK \textgreater$ token representing all unknown words.

\subsection{ConceptNet}\label{conceptNet}
In our experiment, ConceptNet\footnote{https://conceptnet.io. ConceptNet can be Downloaded at http://github.com/commonsense/conceptnet5/wiki/Downloads.} is used as the commonsense knowledge base. Preprocessing of this knowledge base involves removing assertions containing non-English characters or any word outside vocabulary $V$. 1.4M concepts remain. 0.8M concepts are unigrams, 0.43M are bi-grams and the other 0.17M are tri-grams or more. Each concept is associated with an average of 4.3 assertions. More than half of the concepts are associated with only one assertion.

An average of 2.8 concepts can be found in ConceptNet for each message in our Twitter Dialogue Dataset, yielding an average of 150 commonsense assertions (the size of $A_x$). Unsurprisingly, common concepts with more assertions associated are favored in actual human conversations.

It is worth noting that ConceptNet is also noisy due to uncertainties in the constructing process, where 15.5\% of all assertions are considered ``false'' or ``vague'' by human evaluators~\cite{specon}. Our max-pooling strategy used in Tri-LSTM encoder and supervised word embeddings is partly designed to alleviate this weakness. 

\subsection{Parameter Settings}\label{parameter_settings}

In all our models excluding term frequency--inverse document frequency (TF-IDF)~\cite{ramos2003using}, we initialize word embeddings with pretrained GloVe embedding vectors~\cite{pennington2014glove}. The size of hidden units in LSTM models is set to 256 and the word embedding dimension is 100. We use stochastic gradient descent (SGD) for optimizing with batch size of 64. We fixed training rate at 0.001.

\subsection{Results and Analysis}\label{results_and_analysis}
The main results for TF-IDF, word embeddings, memory networks and LSTM models are summarized in Table~\ref{model_performance}. We observe that:

(1) LSTMs perform better at modeling dialogues than word embeddings on our dataset, as shown by the comparison between Tri-LSTM and word embeddings. 

(2) Integrating commonsense knowledge into conversational models boosts model performance, as Tri-LSTM outperforms Dual-LSTM by a certain margin. 

(3) Max-pooling over all commonsense assertions depending on response $y$ is a better method for utilizing commonsense knowledge than attention over memory in our setting, as demonstrated by the gain of performance of word embeddings over memory networks.

\begin{table*}[ht]
\centering
\caption{Model evaluation. $^*$ indicates models with commonsense knowledge integrated. The TF-IDF model is trained following~\cite{lowe2015ubuntu}. The ``Recall@$k$'' method is used for evaluation~\cite{DBLP:journals/corr/LoweSNCP16}. The model is asked to rank a total of $N$ responses containing one positive response and $N-1$ negative responses ($N=10$ according to our test set). If the ranking of the positive response is not larger than $k$, Recall@$k$ is positive for that instance.}
\label{model_performance}
\begin{tabular}{|c|c|c|c|c|c|c|}
\hline
Recall@$k$ & TF-IDF & Word Embeddings$^*$ & Memory Networks$^*$ & Dual-LSTM & Tri-LSTM$^*$   & Human \\ \hline
Recall@1 & 32.6\% & 73.5\% & 72.1\% & 73.6\%  & \textbf{77.5\%} & 87.0\% \\ \hline
Recall@2 & 47.3\% & 84.0\% & 83.6\% & 85.6\%  & \textbf{88.0\%} & -   \\ \hline
Recall@5 & 68.0\% & 95.5\% & 94.2\% & 95.9\%  & \textbf{96.6\%} & -   \\ \hline
\end{tabular}
\end{table*}

\begin{table*}[ht]
\centering
\caption{Case studies for the impact of commonsense assertions. ``Activated Assertion'' is the commonsense assertion entry in $A_x$ chosen by max-pooling. $\diamondsuit$ indicates correct selection. All 4 instances displayed are taken from the test set.}
\label{case_studies}
\begin{tabular}{|c|c|c|}
\hline
Instance & Message                                                                   & \begin{tabular}[c]{@{}c@{}}Response selected \\ by Dual-LSTM\end{tabular}                     \\ \hline
1    & \begin{tabular}[c]{@{}c@{}}i was helping my brother \\ with his chinese.\end{tabular}                            & did yoga help?                                                   \\ \hline
2    & \begin{tabular}[c]{@{}c@{}}bonjour madame, quoi de neuf.
\end{tabular}                                  & yeah me too !                                                   \\ \hline
3    & \begin{tabular}[c]{@{}c@{}}help what colour shoes can i \\ wear with my dress to the wedding?\end{tabular}                & very pale pink or black.$^\diamondsuit$                                              \\ \hline
4    & \begin{tabular}[c]{@{}c@{}}helping mum paint my bedroom.\end{tabular}                                  & \begin{tabular}[c]{@{}c@{}}shouldn't it be your mum \\ helping you? what color are \\ you going for ?$^\diamondsuit$\end{tabular} \\ \hline \hline
Instance & \begin{tabular}[c]{@{}c@{}}Response selected \\ by Tri-LSTM\end{tabular}                                  & \begin{tabular}[c]{@{}c@{}}Activated Assertion\\ (total size of $A_x$)\end{tabular}                  \\ \hline
1    & \begin{tabular}[c]{@{}c@{}}the language sounds interesting! \\ i really gotta learn it !$^\diamondsuit$\end{tabular}                  & \textit{\begin{tabular}[c]{@{}c@{}}chinese, IsA, \\ human\_language (755)\end{tabular}}               \\ \hline
2    & \begin{tabular}[c]{@{}c@{}}loool . you can stick with english ,\\ its all good unless you want\\ to improve your french .$^\diamondsuit$\\ \end{tabular} & \textit{\begin{tabular}[c]{@{}c@{}}bonjour, IsA, \\ hello\_in\_french (9)\end{tabular}}               \\ \hline
3    & very pale pink or black.$^\diamondsuit$                                                           & \textit{\begin{tabular}[c]{@{}c@{}}pink, RelatedTo, \\ colour (1570)\end{tabular}}                 \\ \hline
4    & \begin{tabular}[c]{@{}c@{}}shouldn't it be your mum \\ helping you? what color are \\ you going for ?$^\diamondsuit$\end{tabular}             & \textit{\begin{tabular}[c]{@{}c@{}}paint, RelatedTo, \\ household\_color (959)\end{tabular}}            \\ \hline
\end{tabular}
\end{table*}

We also analyze samples from the test set to gain an insight on how commonsense knowledge supplements the message itself in response selection by comparing Tri-LSTM encoder and Dual-LSTM encoder. 

As illustrated in Table~\ref{case_studies}, instances 1,2 represent cases where commonsense assertions as an external memory module provide certain clues that the other model failed to capture. For example in instance 2, Tri-LSTM selects the response ``...improve your french'' to message ``bonjour madame'' based on a retrieved assertion ``$bonjour, IsA, hello\_in\_french$'', while Dual-LSTM selects an irrelevant response. 
Unsurprisingly, Dual-LSTM is also able to select the correct response in some cases where certain commonsense knowledge is necessary, as illustrated in instance 3. Both models select ``... pink or black'' in response to message ``...what color shoes...'', even though Dual-LSTM does not have access to a helpful assertion ``$pink, RelatedTo,
color$''. 

Informally speaking, such cases suggest that to some extent, Dual-LSTM (models with no memory) is able to encode certain commonsense knowledge in model parameters (e.g., word embeddings) in an implicit way. 
In other cases, e.g., instance 4, the message itself is enough for the selection of the correct response, where both models do equally well.




\section{Conclusion and Future Work}\label{conclusions}

In this paper, we emphasized the role of memory in conversational models. In the open-domain chit-chat setting, we experimented with commonsense knowledge as external memory and proposed to exploit LSTM to encode commonsense assertions to enhance response selection. 

In the other research line of response generation, such knowledge can potentially be used to condition the decoder in favor of more interesting and relevant responses.
Although the gains presented by our new method is not spectacular according to Recall@$k$, our view represents a promising attempt at integrating a large heterogeneous knowledge base that potentially describes the world into conversational models as a memory component. 

Our future work includes extending the commonsense knowledge with common (or factual) knowledge, e.g., to extend the knowledge base coverage by linking more named entities to commonsense knowledge concepts~\cite{cammds}, and developing a better mechanism for utilizing such knowledge instead of the simple max-pooling scheme used in this paper. We would also like to explore the memory of the model for multiple message response pairs in a long conversation.

Lastly, we plan to integrate affective knowledge from SenticNet in the dialogue system in order to enhance its emotional intelligence and, hence, achieve a more human-like interaction. The question, after all, is not whether intelligent machines can have any emotions, but whether machines can be intelligent without any emotions~\cite{minemo}.

\section{Acknowledgements}\label{Acknowledgements}
We gratefully acknowledge the help of Alan Ritter for sharing the twitter dialogue dataset and the NTU PDCC center for providing computing resources.

\bibliography{aaai}

\begin{thebibliography}{}

\bibitem[\protect\citeauthoryear{Bordes and Weston}{2016}]{bordes2016learning}
Bordes, A., and Weston, J.
\newblock 2016.
\newblock Learning end-to-end goal-oriented dialog.
\newblock {\em arXiv preprint arXiv:1605.07683}.

\bibitem[\protect\citeauthoryear{Cambria and Hussain}{2015}]{camsen}
Cambria, E., and Hussain, A.
\newblock 2015.
\newblock {\em Sentic Computing: A Common-Sense-Based Framework for
  Concept-Level Sentiment Analysis}.
\newblock Cham, Switzerland: Springer.

\bibitem[\protect\citeauthoryear{Cambria \bgroup et al\mbox.\egroup
  }{2009}]{camcom}
Cambria, E.; Hussain, A.; Havasi, C.; and Eckl, C.
\newblock 2009.
\newblock Common sense computing: From the society of mind to digital intuition
  and beyond.
\newblock In Fierrez, J.; Ortega, J.; Esposito, A.; Drygajlo, A.; and
  Faundez-Zanuy, M., eds., {\em Biometric ID Management and Multimodal
  Communication}, volume 5707 of {\em Lecture Notes in Computer Science}.
  Berlin Heidelberg: Springer.
\newblock  252--259.

\bibitem[\protect\citeauthoryear{Cambria \bgroup et al\mbox.\egroup
  }{2010}]{campat}
Cambria, E.; Hussain, A.; Durrani, T.; Havasi, C.; Eckl, C.; and Munro, J.
\newblock 2010.
\newblock Sentic computing for patient centered application.
\newblock In {\em {IEEE ICSP}},  1279--1282.

\bibitem[\protect\citeauthoryear{Cambria \bgroup et al\mbox.\egroup
  }{2014}]{cammds}
Cambria, E.; Song, Y.; Wang, H.; and Howard, N.
\newblock 2014.
\newblock Semantic multi-dimensional scaling for open-domain sentiment
  analysis.
\newblock {\em {IEEE} Intelligent Systems} 29(2):44--51.

\bibitem[\protect\citeauthoryear{Cambria \bgroup et al\mbox.\egroup
  }{2016}]{camnt4}
Cambria, E.; Poria, S.; Bajpai, R.; and Schuller, B.
\newblock 2016.
\newblock {SenticNet} 4: A semantic resource for sentiment analysis based on
  conceptual primitives.
\newblock In {\em {COLING}},  2666--2677.

\bibitem[\protect\citeauthoryear{Dodge \bgroup et al\mbox.\egroup
  }{2015}]{dodge2015evaluating}
Dodge, J.; Gane, A.; Zhang, X.; Bordes, A.; Chopra, S.; Miller, A.; Szlam, A.;
  and Weston, J.
\newblock 2015.
\newblock Evaluating prerequisite qualities for learning end-to-end dialog
  systems.
\newblock {\em arXiv preprint arXiv:1511.06931}.

\bibitem[\protect\citeauthoryear{Ghazvininejad \bgroup et al\mbox.\egroup
  }{2017}]{DBLP:journals/corr/GhazvininejadBC17}
Ghazvininejad, M.; Brockett, C.; Chang, M.; Dolan, B.; Gao, J.; Yih, W.; and
  Galley, M.
\newblock 2017.
\newblock A knowledge-grounded neural conversation model.
\newblock {\em CoRR} abs/1702.01932.

\bibitem[\protect\citeauthoryear{Hochreiter and
  Schmidhuber}{1997}]{hochreiter1997long}
Hochreiter, S., and Schmidhuber, J.
\newblock 1997.
\newblock Long short-term memory.
\newblock {\em Neural computation} 9(8):1735--1780.

\bibitem[\protect\citeauthoryear{Li \bgroup et al\mbox.\egroup
  }{2015}]{li2015diversity}
Li, J.; Galley, M.; Brockett, C.; Gao, J.; and Dolan, B.
\newblock 2015.
\newblock A diversity-promoting objective function for neural conversation
  models.
\newblock {\em arXiv preprint arXiv:1510.03055}.

\bibitem[\protect\citeauthoryear{Li \bgroup et al\mbox.\egroup
  }{2016}]{li2016persona}
Li, J.; Galley, M.; Brockett, C.; Spithourakis, G.~P.; Gao, J.; and Dolan, B.
\newblock 2016.
\newblock A persona-based neural conversation model.
\newblock {\em arXiv preprint arXiv:1603.06155}.

\bibitem[\protect\citeauthoryear{Liu \bgroup et al\mbox.\egroup
  }{2016}]{liu2016not}
Liu, C.-W.; Lowe, R.; Serban, I.~V.; Noseworthy, M.; Charlin, L.; and Pineau,
  J.
\newblock 2016.
\newblock How not to evaluate your dialogue system: An empirical study of
  unsupervised evaluation metrics for dialogue response generation.
\newblock {\em arXiv preprint arXiv:1603.08023}.

\bibitem[\protect\citeauthoryear{Lowe \bgroup et al\mbox.\egroup
  }{2015a}]{Lowe-unstructured-text-2015-nips-workshop}
Lowe, R.; Pow, N.; Charlin, L.; Pineau, J.; and Serban, I.~V.
\newblock 2015a.
\newblock Incorporating unstructured textual knowledge sources into neural
  dialogue systems.
\newblock In {\em Machine Learning for Spoken Language Understanding and
  Interaction, NIPS 2015 Workshop}.

\bibitem[\protect\citeauthoryear{Lowe \bgroup et al\mbox.\egroup
  }{2015b}]{lowe2015ubuntu}
Lowe, R.; Pow, N.; Serban, I.; and Pineau, J.
\newblock 2015b.
\newblock The ubuntu dialogue corpus: A large dataset for research in
  unstructured multi-turn dialogue systems.
\newblock {\em arXiv preprint arXiv:1506.08909}.

\bibitem[\protect\citeauthoryear{Lowe \bgroup et al\mbox.\egroup
  }{2016a}]{lowe2016evaluation}
Lowe, R.; Serban, I.~V.; Noseworthy, M.; Charlin, L.; and Pineau, J.
\newblock 2016a.
\newblock On the evaluation of dialogue systems with next utterance
  classification.
\newblock {\em arXiv preprint arXiv:1605.05414}.

\bibitem[\protect\citeauthoryear{Lowe \bgroup et al\mbox.\egroup
  }{2016b}]{DBLP:journals/corr/LoweSNCP16}
Lowe, R.; Serban, I.~V.; Noseworthy, M.; Charlin, L.; and Pineau, J.
\newblock 2016b.
\newblock On the evaluation of dialogue systems with next utterance
  classification.
\newblock {\em CoRR} abs/1605.05414.

\bibitem[\protect\citeauthoryear{Mikolov \bgroup et al\mbox.\egroup
  }{2013}]{mikolov2013efficient}
Mikolov, T.; Chen, K.; Corrado, G.; and Dean, J.
\newblock 2013.
\newblock Efficient estimation of word representations in vector space.
\newblock {\em arXiv preprint arXiv:1301.3781}.

\bibitem[\protect\citeauthoryear{Minsky}{1986}]{minsoc}
Minsky, M.
\newblock 1986.
\newblock {\em The Society of Mind}.
\newblock New York: Simon and Schuster.

\bibitem[\protect\citeauthoryear{Minsky}{2006}]{minemo}
Minsky, M.
\newblock 2006.
\newblock {\em The Emotion Machine: Commonsense Thinking, Artificial
  Intelligence, and the Future of the Human Mind}.
\newblock New York: Simon \& Schuster.

\bibitem[\protect\citeauthoryear{Pennington, Socher, and
  Manning}{2014}]{pennington2014glove}
Pennington, J.; Socher, R.; and Manning, C.~D.
\newblock 2014.
\newblock Glove: Global vectors for word representation.
\newblock In {\em EMNLP}, volume~14,  1532--1543.

\bibitem[\protect\citeauthoryear{Poria \bgroup et al\mbox.\egroup
  }{2015}]{porsent}
Poria, S.; Cambria, E.; Gelbukh, A.; Bisio, F.; and Hussain, A.
\newblock 2015.
\newblock Sentiment data flow analysis by means of dynamic linguistic patterns.
\newblock {\em {IEEE} Computational Intelligence Magazine} 10(4):26--36.

\bibitem[\protect\citeauthoryear{Poria \bgroup et al\mbox.\egroup
  }{2016}]{porlda}
Poria, S.; Chaturvedi, I.; Cambria, E.; and Bisio, F.
\newblock 2016.
\newblock Sentic {LDA}: Improving on {LDA} with semantic similarity for
  aspect-based sentiment analysis.
\newblock In {\em {IJCNN}},  4465--4473.

\bibitem[\protect\citeauthoryear{Rajagopal \bgroup et al\mbox.\egroup
  }{2013}]{rajagopal2013graph}
Rajagopal, D.; Cambria, E.; Olsher, D.; and Kwok, K.
\newblock 2013.
\newblock A graph-based approach to commonsense concept extraction and semantic
  similarity detection.
\newblock In {\em Proceedings of the 22nd International Conference on World
  Wide Web},  565--570.
\newblock ACM.

\bibitem[\protect\citeauthoryear{Ramos and others}{2003}]{ramos2003using}
Ramos, J., et~al.
\newblock 2003.
\newblock Using tf-idf to determine word relevance in document queries.
\newblock In {\em Proceedings of the first instructional conference on machine
  learning}.

\bibitem[\protect\citeauthoryear{Ritter, Cherry, and
  Dolan}{2011}]{ritter2011data}
Ritter, A.; Cherry, C.; and Dolan, W.~B.
\newblock 2011.
\newblock Data-driven response generation in social media.
\newblock In {\em Proceedings of the conference on empirical methods in natural
  language processing},  583--593.
\newblock Association for Computational Linguistics.

\bibitem[\protect\citeauthoryear{Serban \bgroup et al\mbox.\egroup
  }{2016}]{serban2016building}
Serban, I.~V.; Sordoni, A.; Bengio, Y.; Courville, A.; and Pineau, J.
\newblock 2016.
\newblock Building end-to-end dialogue systems using generative hierarchical
  neural network models.
\newblock In {\em Thirtieth AAAI Conference on Artificial Intelligence}.

\bibitem[\protect\citeauthoryear{Singh \bgroup et al\mbox.\egroup
  }{2002}]{singh2002open}
Singh, P.; Lin, T.; Mueller, E.; Lim, G.; Perkins, T.; and Li~Zhu, W.
\newblock 2002.
\newblock Open mind common sense: Knowledge acquisition from the general
  public.
\newblock {\em On the move to meaningful internet systems 2002: CoopIS, DOA,
  and ODBASE}  1223--1237.

\bibitem[\protect\citeauthoryear{Speer and Havasi}{2012}]{specon}
Speer, R., and Havasi, C.
\newblock 2012.
\newblock {ConceptNet} 5: A large semantic network for relational knowledge.
\newblock In Hovy, E.; Johnson, M.; and Hirst, G., eds., {\em Theory and
  Applications of Natural Language Processing}. Springer.
\newblock chapter~6.

\bibitem[\protect\citeauthoryear{Sukhbaatar \bgroup et al\mbox.\egroup
  }{2015}]{sukhbaatar2015end}
Sukhbaatar, S.; Weston, J.; Fergus, R.; et~al.
\newblock 2015.
\newblock End-to-end memory networks.
\newblock In {\em Advances in neural information processing systems},
  2440--2448.

\bibitem[\protect\citeauthoryear{Sutskever, Vinyals, and
  Le}{2014}]{sutskever2014sequence}
Sutskever, I.; Vinyals, O.; and Le, Q.~V.
\newblock 2014.
\newblock Sequence to sequence learning with neural networks.
\newblock In {\em Advances in neural information processing systems},
  3104--3112.

\bibitem[\protect\citeauthoryear{Vinyals and Le}{2015}]{vinyals2015neural}
Vinyals, O., and Le, Q.
\newblock 2015.
\newblock A neural conversational model.
\newblock {\em arXiv preprint arXiv:1506.05869}.

\bibitem[\protect\citeauthoryear{Wang \bgroup et al\mbox.\egroup
  }{2013}]{wancom}
Wang, Q.; Cambria, E.; Liu, C.; and Hussain, A.
\newblock 2013.
\newblock Common sense knowledge for handwritten chinese recognition.
\newblock {\em Cognitive Computation} 5(2):234--242.

\bibitem[\protect\citeauthoryear{Weston \bgroup et al\mbox.\egroup
  }{2015}]{weston2015towards}
Weston, J.; Bordes, A.; Chopra, S.; Rush, A.~M.; van Merri{\"e}nboer, B.;
  Joulin, A.; and Mikolov, T.
\newblock 2015.
\newblock Towards ai-complete question answering: A set of prerequisite toy
  tasks.
\newblock {\em arXiv preprint arXiv:1502.05698}.

\bibitem[\protect\citeauthoryear{Weston, Chopra, and
  Bordes}{2014}]{weston2014memory}
Weston, J.; Chopra, S.; and Bordes, A.
\newblock 2014.
\newblock Memory networks.
\newblock {\em arXiv preprint arXiv:1410.3916}.

\bibitem[\protect\citeauthoryear{Xing \bgroup et al\mbox.\egroup
  }{2016}]{DBLP:journals/corr/XingWWLHZM16}
Xing, C.; Wu, W.; Wu, Y.; Liu, J.; Huang, Y.; Zhou, M.; and Ma, W.
\newblock 2016.
\newblock Topic augmented neural response generation with a joint attention
  mechanism.
\newblock {\em CoRR} abs/1606.08340.

\bibitem[\protect\citeauthoryear{Zhou \bgroup et al\mbox.\egroup
  }{2016}]{zhou2016multi}
Zhou, X.; Dong, D.; Wu, H.; Zhao, S.; Yan, R.; Yu, D.; Liu, X.; and Tian, H.
\newblock 2016.
\newblock Multi-view response selection for human-computer conversation.
\newblock {\em EMNLP'16}.

\end{thebibliography}
\bibliographystyle{aaai}
\end{document}